\documentclass{article}
\usepackage{llm2r_preprint,times}
\usepackage[TS1,OT1]{fontenc}
\usepackage{amsmath,amssymb}
\usepackage{graphicx}
\usepackage{booktabs,multirow,array}
\usepackage[hidelinks,hyperfootnotes=false]{hyperref}
\usepackage{url}

\ifdefined\pdfminorversion
\else
  \AtBeginDocument{}
\fi

\hypersetup{pdftitle={LLM2R: From Large Language Models to 3D Reassemblers},
  pdfsubject={Research preprint},pdfkeywords={3D part assembly, large language models, geometric reassembly}}

\newcommand{\method}{LLM2R}
\newcommand{\bh}{\mathbf{h}}
\newcommand{\bu}{\mathbf{u}}
\newcommand{\bz}{\mathbf{z}}
\newcommand{\bq}{\mathbf{q}}
\newcommand{\bt}{\mathbf{t}}
\newcommand{\bx}{\mathbf{x}}
\newcommand{\R}{\mathbb{R}}

\title{LLM2R: From Large Language Models\\ to 3D Reassemblers}

\author{%
\begin{minipage}{\dimexpr\textwidth-2\tabcolsep\relax}
\centering\normalfont
\textbf{Zhouhan Zhong}\textsuperscript{1}\quad
\textbf{Qiuhong Shen}\textsuperscript{1}\quad
\textbf{Xingyi Yang}\textsuperscript{2}\quad
\textbf{Jinyi Xu}\textsuperscript{1}\quad
\textbf{Xinchao Wang}\textsuperscript{1,\textdagger}\\[0.9ex]
\textsuperscript{1}National University of Singapore\qquad
\textsuperscript{2}The Hong Kong Polytechnic University
\end{minipage}%
}
\hypersetup{pdfauthor={Zhouhan Zhong, Qiuhong Shen, Xingyi Yang, Jinyi Xu, Xinchao Wang}}

\date{}

\begin{document}
\maketitle
{\renewcommand{\thefootnote}{\textdagger}\footnotetext{Corresponding author.}}

\begin{abstract}
3D part assembly aims to reconstruct a complete shape from a set of unassembled parts by predicting a rigid $\mathrm{SE}(3)$ transformation for each part. Because the placement of a part is rarely determined by its own geometry, the task demands reasoning over the entire part set, and existing methods, which rely on pairwise relations or iterative generation and verification, scale poorly to objects with many parts. We present \method{}, which casts part assembly as sequence modeling with a pretrained decoder-only large language model (LLM): a Point-BERT encoder and a Q-Former compress each part into a block of geometric tokens, the blocks of all parts are concatenated, and a causal LLM adapted with Low-Rank Adaptation (LoRA) processes the sequence in one forward pass. Causal attention, however, lets the hidden states of a part observe only the parts that precede it. \method{} therefore appends a block of learnable \emph{summary tokens} after the last part: these tokens follow every part, so their states aggregate the whole assembly at a constant token cost, and a part-aware pose head fuses each part's states with them before predicting its rotation and translation. One architecture thus handles semantic and geometric assembly and predicts all poses in a single non-autoregressive pass. \method{} achieves the highest part accuracy and success rate on all three PartNet categories, the best rotation error, translation error, and part accuracy on PartNeXt, and the lowest translation error and fastest inference on Breaking Bad, where only a fracture-specialized method is more accurate. Its advantage grows with the number of parts: at 100 parts it exceeds GARF by 33.5 points in part accuracy on Breaking Bad and Assembler by 29.1 points on PartNet, and it transfers zero-shot to unseen categories with more than 16 points higher part accuracy than the strongest of three baselines.
\end{abstract}

\section{Introduction}
\label{sec:intro}

When humans are presented with the scattered components of a disassembled object, such as furniture parts, archaeological fragments, or shattered everyday items, they can often infer the target structure and a spatially consistent arrangement of every component from local geometric cues and structural priors. This capacity is a hallmark of \emph{spatial intelligence}~\citep{lu_survey_2025-1}, and computationally it manifests as the \emph{3D part assembly problem}: given a set of unassembled parts, predict the $\mathrm{SE}(3)$ rigid transformation of each part that reconstructs the complete shape. The task is relevant to manufacturing, robotics, archaeology, and restoration, often without a CAD model or template, so a consistent configuration must be inferred from the part geometry alone. What makes it hard is that the placement of a part is rarely determined by the part itself; every pose depends on the whole set.

Existing methods exhibit three persistent limitations~\citep{zhan2020generative,zhang_3d_2022,du_generative_2024,lee20243d,wang2024puzzlefusion++}. First, most approaches model inter-part relationships through pairwise matching or local compatibility scores, which makes it difficult to capture higher-order constraints and to enforce global consistency across all parts~\citep{lee20243d,lee2025combinative,wu2023leveraging}. Second, many methods depend on auxiliary cues such as reference images, CAD templates, or category-specific priors that are unavailable in open-domain settings~\citep{li2020learning,zhang2025manual,lu2410jigsaw++}. Third, they scale poorly with the part count: long-range dependencies among many components are hard to model without long-context support, and the cost of iterative candidate generation and verification grows rapidly~\citep{narayan2022rgl,xu2025spaformer,zhao2025assembler,zhang2024scalable}.

Our starting point is that this is a sequence modeling problem of the kind that decoder-only large language models (LLMs) already solve at scale: every output must be consistent with a long context of tokens, and pretrained decoders bring long-context attention, efficient key-value caching, and weights that can be adapted with few trainable parameters. We therefore propose \textbf{\method{}}, which treats each part as a block of geometric tokens, produced by a Point-BERT encoder~\citep{yu2022point} and a Q-Former~\citep{li2023blip}, concatenates the blocks of all parts, and processes the sequence with a pretrained decoder-only LLM~\citep{yang2025qwen3} adapted through Low-Rank Adaptation (LoRA)~\citep{hu2022lora}. The input is point clouds only, without images, templates, or category-specific priors, and the same architecture handles semantic part assembly and geometric fractured reassembly.

A decoder-only LLM is causal, so the hidden states of the $i$-th part attend to the instruction and to parts $1,\ldots,i$ but not to the parts that follow: a pose read from a part's own tokens would place the first leg of a chair without having seen the seat or the other legs. Rather than altering the attention mask, so that the pretrained decoder is reused unchanged, \method{} appends a block of $S$ learnable \emph{summary tokens} after the last part. These tokens follow every part, so their contextual states are computed with access to the complete assembly, once per object, within the same forward pass, and at a cost that is constant in the number of parts. A part-aware pose head then fuses the hidden states of each part with the summary states and regresses its rotation and translation, so that all poses are predicted in parallel from one pass over the sequence.

We evaluate \method{} on PartNet~\citep{mo2019partnet}, Breaking Bad~\citep{sellan2022breaking}, and the recently released PartNeXt benchmark~\citep{wang2025partnext}, covering both semantic and geometric assembly, and study zero-shot transfer to unseen categories, scaling with the number of parts, and the contribution of each component. Our contributions are as follows.
\begin{itemize}
\item We cast 3D part assembly as causal sequence modeling over geometric part tokens and adapt a pretrained decoder-only LLM to it with LoRA, yielding a single point-cloud-only framework for semantic and geometric assembly that predicts all poses in one non-autoregressive pass.
\item We introduce a block of learnable summary tokens and a part--summary fusion head that condition every pose on the whole assembly at a constant token cost without modifying the causal decoder; ablations identify the summary block as the component that matters most for part accuracy.
\item \method{} achieves the best part accuracy on PartNet and PartNeXt and the lowest translation error and inference time on Breaking Bad. Its advantage grows with the part count, reaching 33.5 and 29.1 part-accuracy points over GARF and Assembler at 100 parts, and it transfers zero-shot to unseen categories with more than 16 points higher part accuracy than the strongest of three baselines.
\end{itemize}
To our knowledge, \method{} is the first method to adapt a pretrained decoder-only LLM to direct $\mathrm{SE}(3)$ pose prediction for 3D part assembly.

\section{Related Work}
\label{sec:related}

\paragraph{3D shape assembly.}
\label{sec:related-work-assembly}
\emph{Semantic part assembly} predicts the poses of annotated parts by modeling inter-part relations and refining configurations iteratively~\citep{zhan2020generative,narayan2022rgl}; this line has evolved toward sequence-aware formulations~\citep{xu2025spaformer}, hierarchical formulations such as 3DHPA~\citep{du_generative_2024} and CFPA~\citep{zhang2025cfpa}, and distribution learning via diffusion-style processes that sample multiple plausible assemblies and support variable part counts~\citep{cheng_score-pa_2023,zhao2025assembler}. \emph{Geometric fractured reassembly} targets unlabeled fragments for which fracture-interface complementarity is central. Benchmarks~\citep{sellan2022breaking} have catalyzed agglomerative denoise-and-verify pipelines~\citep{wang2024puzzlefusion++}, $\mathrm{SE}(3)$-equivariant representations~\citep{wu2023leveraging}, explicit matching of surface similarity and volumetric complementarity~\citep{lee2025combinative,lee20243d}, flow- and diffusion-based shape priors~\citep{scarpellini2024diffassemble,lu2410jigsaw++,li2025garf,sun_rectified_2025}, and generation-assisted formulations such as CRAG~\citep{jiang2026crag}. BrickGPT applies autoregressive language modeling to brick-based construction~\citep{pun2025brickgpt} rather than to the reassembly of given parts. \method{} instead processes all parts in a single pass of a long-context LLM and conditions each pose on a learned summary of the whole assembly.

\paragraph{LLMs for point cloud understanding.}
A growing body of work encodes point clouds into tokens that are concatenated with text so that a decoder-only LLM acquires 3D-oriented dialogue and reasoning capabilities. PointLLM~\citep{xu2024pointllm}, Point-Bind~\citep{guo2023point}, ShapeLLM~\citep{qi2024shapellm}, and ShapeLLM-Omni~\citep{ye2025shapellm} address object-level description, question answering, and interaction, while 3D-LLM~\citep{hong20233d}, Chat-3D~\citep{wang2023chat}, LSceneLLM~\citep{zhi2025lscenellm}, and MiniGPT-3D~\citep{tang2024minigpt} extend the interface to scenes. These methods generate free-form text; \method{} instead uses a causal LLM~\citep{yang2025qwen3} to predict per-part $\mathrm{SE}(3)$ transformations and equips it with a summary block so that each prediction can depend on the entire part set.

\paragraph{Pretrained language models for continuous outputs.}
Decoder-only language models have also been adapted to tasks whose outputs are not text. Frozen pretrained Transformers transfer to sequence tasks in other modalities when only a small fraction of their parameters is trained~\citep{lu2022frozen}; vision-language-action models such as RT-2~\citep{brohan2023rt2} and OpenVLA~\citep{kim2024openvla} fine-tune a pretrained vision-language model to emit robot actions discretized into tokens; and ChatPose~\citep{feng2024chatpose} regresses 3D human pose parameters from the hidden state of a dedicated token. \method{} follows this line in reusing a pretrained causal decoder with LoRA, but it reads out continuous $\mathrm{SE}(3)$ poses for a variable number of parts in one pass, each conditioned on summary states of the whole input sequence.

\paragraph{Auxiliary tokens in Transformers.}
Content-free tokens that provide extra positions for computation or aggregation appear as global tokens in sparse attention~\citep{zaheer2020bigbird}, memory tokens~\citep{burtsev2020memory}, registers in vision Transformers~\citep{darcet2024registers}, and latent queries in Perceiver~\citep{jaegle2021perceiver} and the Q-Former of BLIP-2~\citep{li2023blip}. The summary block of \method{} differs in placement and use. It is appended \emph{after} all inputs of a causal decoder, so its states summarize the complete part sequence, and these states are fused position-wise with each part's states inside the pose head rather than consumed by a language model.

\paragraph{Point cloud encoders.}
PointNet and its successors process unordered point sets with shared MLPs, symmetric aggregation, hierarchical grouping, dynamic graphs, point convolutions, or point-cloud self-attention~\citep{qi2017pointnet,qi2017pointnet++,wang2019dynamic,thomas2019kpconv,zhao2021point}, and self-supervised pretraining by contrastive learning~\citep{xie2020pointcontrast} or masked point modeling~\citep{yu2022point,pang2023masked} yields transferable representations. We adopt a pretrained Point-BERT encoder and learn a Q-Former that compresses its patch features into a fixed number of tokens in the language model's embedding space.

\section{Method}
\label{sec:method}

\begin{figure}[t]
\begin{center}
\makebox[\linewidth][c]{\includegraphics[width=1.15\linewidth]{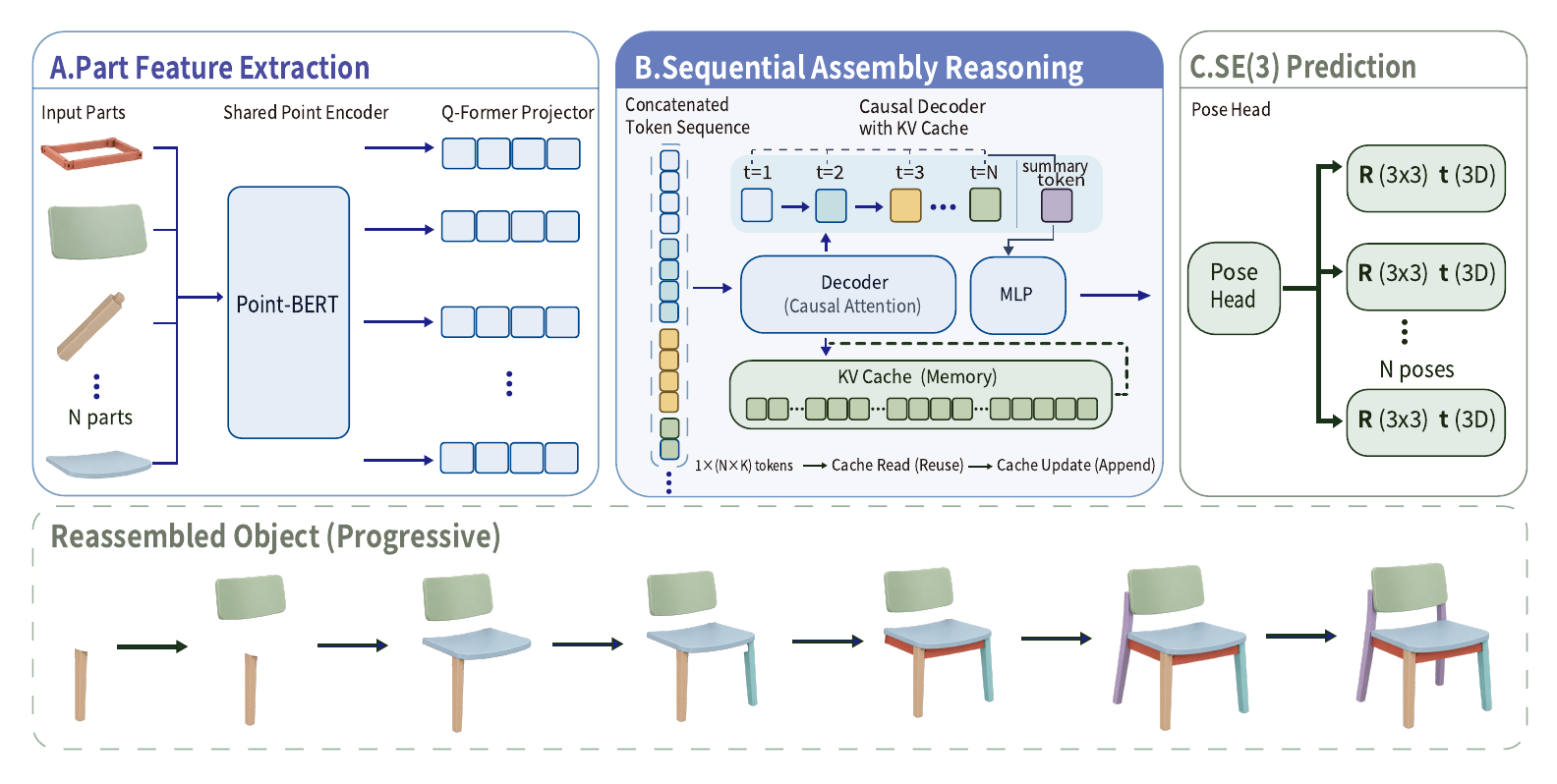}}
\end{center}
\caption{Overview of \method{}. (A) Each part is encoded by a shared Point-BERT encoder and compressed by a Q-Former into $K$ geometric tokens. (B) The token blocks of all $N$ parts are concatenated and followed by the learnable summary block; a causal decoder processes the sequence in a single forward pass, so the summary positions attend to every part, and the key-value cache of the parts processed so far can be reused when further parts are appended. (C) A pose head fuses each part's states with the summary states and predicts the rotation $R$ and translation $\mathbf t$ of every part.}
\label{fig:pipeline}
\end{figure}

\method{} takes a set of unassembled part point clouds as input and predicts a rigid $\mathrm{SE}(3)$ transformation for each part that places it in the assembled shape (Figure~\ref{fig:pipeline}).

\paragraph{Problem statement.}
Let $\mathcal{P}=\{P_1,\ldots,P_N\}$ denote $N$ input parts, where $P_i\in\R^{n_i\times3}$ is the point set of part $i$ expressed in its canonical frame. The goal is to estimate a pose $T_i=(R_i,\bt_i)\in\mathrm{SE}(3)$ for every part such that $R_iP_i+\bt_i$ reproduces its placement in the global frame of the assembled object. Rotations are represented as unit quaternions $\bq_i\in\mathbb{S}^3$ (scalar first), with $R_i=R(\bq_i)$. Canonical frames and pose targets are derived from the part geometry itself (Appendix~\ref{sec:data_processing}); no images, templates, or semantic labels are used.

\subsection{Geometric Tokenization and Sequence Construction}
\label{sec:assembler}

\paragraph{Part feature encoding.}
Each part is encoded independently by a pretrained Point-BERT encoder $\Phi$~\citep{yu2022point}, which partitions $P_i$ into $G$ local neighborhoods by farthest point sampling and $k$-nearest-neighbor grouping, embeds each neighborhood with a small point network, and processes the resulting patch tokens with a Transformer. We use its patch-level output, one $C$-dimensional descriptor per neighborhood, $F_i=\Phi(P_i)\in\R^{G\times C}$. The encoder is initialized from masked point modeling on ShapeNet and kept frozen for most of training (Section~\ref{sec:setup}); $G=64$, each neighborhood contains 32 points, and $C=8192$ is the size of the discrete point vocabulary of Point-BERT, whose logits form the patch descriptor.

\paragraph{Compression into the LLM embedding space.}
A Q-Former~\citep{li2023blip} compresses the $G$ patch descriptors of a part into $K$ tokens in the hidden dimension $d$ of the language model. The descriptors are first projected and normalized, $\tilde F_i=\mathrm{LN}(W_{\mathrm{in}}F_i+\mathbf b_{\mathrm{in}})\in\R^{G\times d}$. A set of $K$ learnable queries $Q^{(0)}\in\R^{K\times d}$ is then refined by $N_\ell=2$ pre-norm layers; each layer applies self-attention among the queries, cross-attention from the queries to $\tilde F_i$, and a feed-forward sublayer, all with residual connections:
\begin{equation}
Q^{(\ell)}=\mathrm{QFormerLayer}\bigl(Q^{(\ell-1)},\tilde F_i\bigr),\qquad
V_i=\mathrm{LN}\bigl(Q^{(N_\ell)}\bigr)\in\R^{K\times d}.
\label{eq:projector}
\end{equation}
The queries are shared across parts, so each part is represented by a block of $K$ geometric tokens in a fixed order; we use $K=64$ (a per-patch MLP projector is compared in Section~\ref{sec:ablation}).

\paragraph{Summary tokens.}
We introduce a block of $S$ learnable summary tokens $S^{(0)}=[\mathbf s^{(0)}_1;\ldots;\mathbf s^{(0)}_S]\in\R^{S\times d}$, initialized from $\mathcal N(0,0.02^2)$, shared by all objects, trained jointly with the pose head, and appended once per object after the last part. The block carries no geometric content, but its positions follow every part, so that their hidden states can summarize the entire assembly (Section~\ref{sec:summary}); we set $S=K=64$ so that the summary block can be fused with a part block position by position (Section~\ref{sec:head_and_loss}).

\paragraph{Decoder input.}
The geometric tokens and the summary block are inserted into a chat-formatted instruction, inside the user message immediately after the instruction text. Let $E_{\mathrm{pre}}$ and $E_{\mathrm{post}}$ denote the token embeddings of the chat template before and after the inserted blocks (system message and instruction; end-of-turn token and assistant header). The decoder input is
\begin{equation}
X=\bigl[E_{\mathrm{pre}};\,V_1;\,V_2;\,\ldots;\,V_N;\,S^{(0)};\,E_{\mathrm{post}}\bigr]\in\R^{L\times d},
\qquad L=L_{\mathrm{pre}}+NK+S+L_{\mathrm{post}}.
\label{eq:sequence}
\end{equation}
The index sets $\mathcal I_i$ and $\mathcal I_S$ of the part and summary blocks are recorded for readout; parts are presented in the dataset order, without any ordering heuristic.

\subsection{Causal Decoding with Assembly-Wide Summary States}
\label{sec:summary}

\paragraph{Causal decoder.}
The sequence $X$ is processed by a pretrained decoder-only LLM $\mathcal D_\theta$,
\begin{equation}
H=\mathcal D_\theta(X)\in\R^{L\times d},\qquad
H_i=H[\mathcal I_i]\in\R^{K\times d},\qquad
U=H[\mathcal I_S]\in\R^{S\times d},
\label{eq:decoder}
\end{equation}
where $H_i=(\bh_{i1},\ldots,\bh_{iK})$ are the final-layer states of the tokens of part $i$ and $U=(\bu_1,\ldots,\bu_S)$ are those of the summary block. The decoder is causal: at every attention layer a token attends only to itself and to preceding positions, with padding positions masked as well. We use the language model of Qwen3.5-0.8B~\citep{yang2025qwen3}, a hybrid decoder with $d=1024$ and 24 layers in which three gated linear-attention layers~\citep{yang2025gated} are followed by one full softmax-attention layer, both causal.

\paragraph{Information flow.}
The state $\bh_{ij}$ of the $j$-th token of part $i$ depends only on the instruction, the blocks $V_1,\ldots,V_{i-1}$, and the first $j$ tokens of $V_i$, whereas each summary state $\bu_j$ follows all $NK$ geometric tokens and depends on the complete part set:
\begin{equation}
\bh_{ij}=f_\theta\bigl(E_{\mathrm{pre}},\,V_{1:i-1},\,V_i[1{:}j]\bigr),\qquad
\bu_j=f_\theta\bigl(E_{\mathrm{pre}},\,V_{1:N},\,S^{(0)}[1{:}j]\bigr).
\label{eq:flow}
\end{equation}
Reading the pose of part $i$ from $H_i$ alone, as in a last-token readout, would thus condition early parts on a strict prefix of the assembly. The summary block resolves this asymmetry without modifying the attention mask: the whole assembly becomes available to every part through $U$, which is computed once per object at a cost that is constant in the number of parts. The model is not invariant to the part order, since reordering changes the prefixes seen by the part states and the summary states (Section~\ref{sec:ablation}).

\paragraph{Parameter-efficient adaptation.}
The decoder is adapted with LoRA~\citep{hu2022lora}, which replaces each frozen weight $W_0$ of the query, key, value, and output projections by $W_0+\tfrac{\alpha}{r}BA$ with trainable low-rank factors $A\in\R^{r\times d_{\mathrm{in}}}$ and $B\in\R^{d_{\mathrm{out}}\times r}$. The base LLM remains frozen; the trainable parameters are the LoRA factors, the Q-Former, the summary block $S^{(0)}$, and the pose head, with the point encoder unfrozen only in a final fine-tuning stage (Section~\ref{sec:setup}).

\subsection{Part-Aware Pose Prediction}
\label{sec:head_and_loss}

\paragraph{Part--summary fusion.}
For each part $i$, the head pairs the $j$-th part state with the $j$-th summary state, fuses each pair with a shared linear layer, and averages over the block:
\begin{equation}
\mathbf f_{ij}=\mathrm{GELU}\bigl(W_f[\bh_{ij};\bu_j]+\mathbf b_f\bigr),\qquad
\bz_i=\frac1K\sum_{j=1}^{K}\mathbf f_{ij},\qquad W_f\in\R^{d\times 2d}.
\label{eq:fusion}
\end{equation}
Since $S=K$, the pairing is one-to-one (summary states missing after context truncation would be replaced by zeros, which does not occur within the context limit used in our experiments), and the pooled feature $\bz_i$ is shared by the two pose branches below.

\paragraph{Rotation prediction.}
Rotation is predicted through a fixed set of $N_a=32$ anchor quaternions $\{\mathbf c_m\}_{m=1}^{N_a}$ (the identity and axis-aligned $90^\circ$ and $180^\circ$ rotations with both quaternion signs; Appendix~\ref{sec:anchors}). A rotation MLP $g_{\mathrm{rot}}$ maps $\bz_i$ to $N_a$ logits, which are normalized by a softmax and used to form a convex combination of the anchors:
\begin{equation}
\mathbf a_i=g_{\mathrm{rot}}(\bz_i)\in\R^{N_a},\qquad
\pi_i=\mathrm{softmax}(\mathbf a_i),\qquad
\hat\bq_i=\frac{\sum_{m=1}^{N_a}\pi_{im}\,\mathbf c_m}{\bigl\|\sum_{m=1}^{N_a}\pi_{im}\,\mathbf c_m\bigr\|_2}.
\label{eq:rotation}
\end{equation}
Because the anchors positively span $\R^4$, every unit quaternion is reachable after normalization, so Eq.~\eqref{eq:rotation} is a continuous parameterization of $\mathrm{SO}(3)$ rather than a classification over discrete rotations. A second MLP regresses the translation directly, $\hat\bt_i=g_{\mathrm{trans}}(\bz_i)\in\R^3$; both $g_{\mathrm{rot}}$ and $g_{\mathrm{trans}}$ are three-layer perceptrons with GELU activations and dropout, and the rotation branch applies layer normalization to its logits before the softmax.

\paragraph{Training objectives.}
Let $(\bq_i^\star,\bt_i^\star)$ be the target pose of part $i$. Since $\bq$ and $-\bq$ encode the same rotation, the rotation loss takes the smaller of the two squared distances on the quaternion sphere, and the translation loss is the mean squared coordinate error:
\begin{equation}
\mathcal L_{\mathrm{pose}}=\frac1N\sum_{i=1}^{N}\Bigl[
\lambda_R\,\min_{\sigma\in\{-1,+1\}}\bigl\|\hat\bq_i-\sigma\bq_i^\star\bigr\|_2^2
+\lambda_t\,\tfrac13\bigl\|\hat\bt_i-\bt_i^\star\bigr\|_2^2\Bigr].
\label{eq:pose_loss}
\end{equation}
To supervise the assembled geometry directly, a Chamfer term transforms the part by the predicted and by the target pose and compares the two point sets,
\begin{align}
\mathcal L_{\mathrm{CD}}&=\frac1N\sum_{i=1}^{N}\mathrm{CD}\bigl(\hat R_iP_i+\hat\bt_i,\;R_i^\star P_i+\bt_i^\star\bigr),
\label{eq:chamfer}\\
\mathrm{CD}(\mathcal X,\mathcal Y)&=\frac12\Bigl[\frac{1}{|\mathcal X|}\sum_{\bx\in\mathcal X}\min_{\mathbf y\in\mathcal Y}\|\bx-\mathbf y\|_2^2+\frac{1}{|\mathcal Y|}\sum_{\mathbf y\in\mathcal Y}\min_{\bx\in\mathcal X}\|\mathbf y-\bx\|_2^2\Bigr],
\nonumber
\end{align}
computed on 1{,}024 random points per part; unlike the quaternion loss, it is invariant to the sign and axis ambiguities of the canonical frames of symmetric parts, because it compares transformed geometry rather than pose parameters.

\paragraph{Interchangeable parts.}
Objects frequently contain geometrically identical parts, such as the four legs of a chair, whose correspondence to individual pose targets is arbitrary, and fixed assignments bias a regressor toward the mean of the group. We therefore group interchangeable parts by a geometric signature (sorted bounding-box extents and covariance spectrum). A repulsion term then penalizes predicted pairwise distances within a group $\mathcal G$ that fall below the target distances,
\begin{equation}
\mathcal L_{\mathrm{rep}}=\operatorname*{mean}_{\mathcal G}\;\operatorname*{mean}_{a<b\in\mathcal G}\;
\max\bigl(0,\;\|\bt_a^\star-\bt_b^\star\|_2-\|\hat\bt_a-\hat\bt_b\|_2\bigr)^2.
\label{eq:repulsion}
\end{equation}
The overall objective,
\begin{equation}
\mathcal L=\mathcal L_{\mathrm{pose}}+\lambda_{\mathrm{CD}}\,\mathcal L_{\mathrm{CD}}+\lambda_{\mathrm{rep}}\,\mathcal L_{\mathrm{rep}},
\label{eq:total_loss}
\end{equation}
is computed per object and averaged over the mini-batch; no language-modeling loss is used, so the summary tokens are shaped solely by the pose objectives. At test time the sequence in Eq.~\eqref{eq:sequence} is decoded once and all poses are read out in parallel from $\{H_i\}_{i=1}^N$ and $U$; no pose is fed back and no refinement is performed, so inference costs one forward pass over $L$ tokens. Because the decoder is causal, the key-value cache of the instruction and of the parts processed so far remains valid when further part blocks are appended, so an assembly can also be processed incrementally, re-evaluating only the new part block and the summary block as each part arrives (Section~\ref{sec:scalability}).

\section{Experiments}
\label{sec:experiments}

\subsection{Experimental Setup}
\label{sec:setup}

\paragraph{Benchmarks and protocols.}
We evaluate on three benchmarks and, for each, adopt the training and evaluation protocol of the strongest published baseline, so that the reported numbers are directly comparable. Breaking Bad~\citep{sellan2022breaking} defines components as irregular fragments produced by physically simulated fracture and thus poses a geometric reassembly problem without semantic annotation. Following GARF~\citep{li2025garf}, we train one model on all Breaking Bad subsets and evaluate it on the volume-constrained Everyday (7,872 assemblies) and Artifact (3,697 assemblies) test sets. PartNet~\citep{mo2019partnet} provides fine-grained semantic parts; following Assembler~\citep{zhao2025assembler}, we train a second model on the Chair, Table, and Lamp categories with the official splits. PartNeXt~\citep{wang2025partnext} provides 23,519 textured meshes with fine-grained, hierarchical part annotations; following CRAG~\citep{jiang2026crag}, we train a third model on its training split of 15,563 shapes and evaluate on 3,903 shapes with 2 to 20 parts, extracting one point set per annotated part (Appendix~\ref{sec:data_processing}).

\paragraph{Metrics.}
On Breaking Bad, we follow GARF~\citep{li2025garf} and report the root mean square rotation error RMSE(R) in degrees and translation error RMSE(T), Part Accuracy (PA), the percentage of parts whose Chamfer distance to the ground-truth placement lies below 0.01, and the Chamfer Distance (CD) between the assembled object and the ground-truth shape. On PartNet, we follow Assembler~\citep{zhao2025assembler} and additionally report Connectivity Accuracy (CA), whether adjacent parts are correctly connected, and Success Rate (SR), the percentage of assemblies in which all parts satisfy the PA criterion. On PartNeXt, we follow CRAG~\citep{jiang2026crag}, which adopts the metric definitions of GARF, and report the rotation error (RE) in degrees, the translation RMSE, PA, and CD averaged over all test shapes.

\paragraph{Implementation details.}
The backbone is the Qwen3.5-0.8B language model (24 layers, $d=1024$) with LoRA ($r=32$, $\alpha=64$) on the query, key, value, and output projections of its attention layers; the point encoder is Point-BERT pretrained on ShapeNet ($G=64$ neighborhoods of 32 points), and $K=S=64$. We train for 100 epochs (about 10 hours) on 8 NVIDIA H200 GPUs with a batch size of 32, using AdamW (learning rate $10^{-3}$) in bfloat16 with a context limit of 8{,}192 tokens, and select the checkpoint with the lowest validation loss. The point encoder is frozen for the first 80 epochs, during which the 42.2M trainable parameters comprise the LoRA matrices, the Q-Former, the summary block, and the pose head, and is fine-tuned jointly for the remaining 20 epochs. All loss weights are one, $\lambda_R=\lambda_t=\lambda_{\mathrm{CD}}=\lambda_{\mathrm{rep}}=1$; the remaining settings are listed in Appendix~\ref{sec:implementation_appendix}.

\subsection{Benchmark Comparisons}
\label{sec:benchmarks}

\paragraph{Breaking Bad.}
Table~\ref{tab:breaking-bad} compares \method{} with representative fractured-reassembly methods on the volume-constrained Everyday and Artifact subsets under the protocol of GARF. \method{} achieves the lowest translation error on both subsets (RMSE(T) of 0.46 and 0.53, against 1.34 and 1.77 for GARF) and the lowest inference time (0.092\,s per instance, against 0.225\,s for GARF and 0.185\,s for Jigsaw), owing to its single non-autoregressive pass. GARF remains ahead on rotation error, PA, and CD: its rotation error is about half that of \method{} on both subsets, its CD is lower by a factor of ten on Everyday and three on Artifact, and the PA gap is 2.2 and 3.2 points (94.77\% and 93.34\% against 92.57\% and 90.11\%). GARF is pretrained on fracture surfaces and specialized to this setting, whereas \method{} is a unified architecture without fracture-specific priors; it outperforms all remaining methods on every metric. Qualitative comparisons are given in Appendix~\ref{sec:qualitative_appendix}.

\begin{table}[t]
\caption{Quantitative comparison on the volume-constrained Breaking Bad dataset~\citep{sellan2022breaking}. RMSE(R) in degrees, RMSE(T) in $\times10^{-2}$, PA in percent, CD in $\times10^{-3}$, IT in seconds per instance. Baseline results follow the protocol of GARF. The best result per column is in bold and the second best is underlined.}
\label{tab:breaking-bad}
\begin{center}
\scriptsize\renewcommand{\arraystretch}{0.92}
\setlength{\tabcolsep}{3.5pt}
\begin{tabular}{l ccccc ccccc}
\toprule
& \multicolumn{5}{c}{\textbf{Everyday}} & \multicolumn{5}{c}{\textbf{Artifact}} \\
\cmidrule(lr){2-6}\cmidrule(lr){7-11}
\textbf{Method} & RMSE(R)$\downarrow$ & RMSE(T)$\downarrow$ & PA$\uparrow$ & CD$\downarrow$ & IT$\downarrow$
                & RMSE(R)$\downarrow$ & RMSE(T)$\downarrow$ & PA$\uparrow$ & CD$\downarrow$ & IT$\downarrow$ \\
\midrule
Global~\citep{li2020partgen} & 80.50 & 14.60 & 28.70 & 13.00 & -- & -- & -- & -- & -- & -- \\
LSTM~\citep{wu2020pqnet} & 82.70 & 15.10 & 27.50 & 13.30 & -- & -- & -- & -- & -- & -- \\
DGL~\citep{zhan2020generative} & 80.30 & 13.90 & 31.60 & 11.80 & -- & -- & -- & -- & -- & -- \\
Jigsaw~\citep{lu2023jigsaw} & 42.19 & 6.85 & 68.89 & 8.22 & \underline{0.185} & 43.75 & 7.91 & 65.12 & 8.50 & \underline{0.171} \\
PMTR~\citep{lee20243d} & 31.57 & 9.95 & 70.60 & 5.56 & -- & -- & -- & -- & -- & -- \\
PF++~\citep{wang2024puzzlefusion++} & 35.61 & 6.05 & 76.17 & 2.78 & 0.385 & 47.03 & 10.63 & 57.97 & 8.24 & 0.379 \\
GARF~\citep{li2025garf} & \textbf{6.68} & \underline{1.34} & \textbf{94.77} & \textbf{0.25} & 0.225 & \textbf{7.67} & \underline{1.77} & \textbf{93.34} & \textbf{0.81} & 0.239 \\
\textbf{\method{} (ours)} & \underline{14.35} & \textbf{0.46} & \underline{92.57} & \underline{2.43} & \textbf{0.092} & \underline{16.81} & \textbf{0.53} & \underline{90.11} & \underline{2.73} & \textbf{0.094} \\
\bottomrule
\end{tabular}
\end{center}
\end{table}

\paragraph{PartNet.}
Table~\ref{tab:partnet-main} reports the comparison on the Chair, Table, and Lamp categories of PartNet under the protocol of Assembler, including the hierarchical methods 3DHPA~\citep{du_generative_2024} and CFPA~\citep{zhang2025cfpa}. \method{} attains the highest PA on all three categories (2.2, 2.2, and 8.3 points above the strongest baseline) and the highest SR among the methods that report it, with a large margin on Chair (41.40\% against 22.59\% for Assembler); its CA is the best on Chair and second or third on Table and Lamp. CD is the metric on which the hierarchical methods remain ahead: 3DHPA and CFPA obtain lower CD on every category, and on Table and Lamp \method{} trails several baselines. This gap recurs on PartNeXt, and we discuss it there.

\begin{table}[t]
\caption{Quantitative comparison on PartNet~\citep{mo2019partnet} (Chair, Table, and Lamp). CD in $\times10^{-3}$; PA, CA, and SR in percent. Baseline results follow the protocol of Assembler; the 3DHPA and CFPA rows are taken from the respective papers, which do not report SR. The best result per column is in bold and the second best is underlined.}
\label{tab:partnet-main}
\begin{center}
\scriptsize\renewcommand{\arraystretch}{0.92}
\setlength{\tabcolsep}{3.5pt}
\begin{tabular}{l cccc cccc cccc}
\toprule
& \multicolumn{4}{c}{\textbf{Chair}} & \multicolumn{4}{c}{\textbf{Table}} & \multicolumn{4}{c}{\textbf{Lamp}} \\
\cmidrule(lr){2-5}\cmidrule(lr){6-9}\cmidrule(lr){10-13}
\textbf{Method} & CD$\downarrow$ & PA$\uparrow$ & CA$\uparrow$ & SR$\uparrow$ & CD$\downarrow$ & PA$\uparrow$ & CA$\uparrow$ & SR$\uparrow$ & CD$\downarrow$ & PA$\uparrow$ & CA$\uparrow$ & SR$\uparrow$ \\
\midrule
DGL~\citep{zhan2020generative} & 9.60 & 37.50 & 24.26 & 7.60 & -- & -- & -- & -- & 7.82 & 32.33 & 42.06 & 12.38 \\
RGL~\citep{narayan2022rgl} & 10.23 & 43.82 & 25.97 & 7.43 & -- & -- & -- & -- & 15.24 & 34.19 & 47.18 & 14.86 \\
Score-PA~\citep{cheng_score-pa_2023} & 7.41 & 41.74 & 28.38 & 5.06 & 4.54 & 51.57 & 39.85 & 7.53 & 8.68 & 31.05 & 48.12 & 8.59 \\
SPAFormer~\citep{xu2025spaformer} & 7.02 & 55.48 & 39.16 & 13.68 & 3.93 & 64.14 & \textbf{61.90} & 33.98 & 12.80 & 36.76 & 45.66 & 15.54 \\
Assembler~\citep{zhao2025assembler} & 9.20 & 61.85 & \underline{49.82} & \underline{22.59} & 8.02 & 65.04 & 56.15 & \underline{39.07} & 26.88 & \underline{44.04} & 58.16 & \underline{27.88} \\
3DHPA~\citep{du_generative_2024} & \underline{5.10} & 64.13 & 49.28 & -- & \textbf{2.80} & 64.83 & 58.45 & -- & \textbf{7.00} & 38.80 & \textbf{64.16} & -- \\
CFPA~\citep{zhang2025cfpa} & \textbf{4.90} & \underline{69.24} & 49.20 & -- & \underline{3.30} & \underline{68.48} & 58.51 & -- & \underline{7.70} & 36.35 & \underline{63.32} & -- \\
\textbf{\method{} (ours)} & 6.21 & \textbf{71.42} & \textbf{53.87} & \textbf{41.40} & 5.47 & \textbf{70.67} & \underline{60.58} & \textbf{42.09} & 17.14 & \textbf{52.31} & 60.83 & \textbf{30.09} \\
\bottomrule
\end{tabular}
\end{center}
\end{table}

\paragraph{PartNeXt.}
\label{sec:partnext}
We further evaluate \method{} on PartNeXt against four recent baselines: GARF~\citep{li2025garf}, Rectified Point Flow (RPF)~\citep{sun_rectified_2025}, CRAG~\citep{jiang2026crag}, and Assembler~\citep{zhao2025assembler}. As reported in Table~\ref{tab:partnext}, \method{} ranks first on all three pose-centric metrics, with the lowest rotation error (41.12 against 45.12 for CRAG), the lowest translation RMSE (5.24 against 9.13), and the highest PA (67.93\% against 65.89\%), but trails on CD (6.37 against 2.40 for CRAG and 3.10 for GARF). The two groups of metrics measure different things: RE, RMSE, and PA measure how correctly each part is posed, whereas CD rewards tight surface contact and is dominated by the parts placed far from the surface. We hypothesize that the fracture-oriented pretraining of GARF and the shape prior of CRAG favor locally well-fitted interfaces, which lowers CD even when more parts are misplaced, whereas the few parts that \method{} misplaces lie far from the surface and inflate CD; the same pattern would explain the CD gap on PartNet.

\begin{table}[t]
\begin{minipage}[t]{0.5\linewidth}
\caption{Quantitative comparison on PartNeXt~\citep{wang2025partnext}. RE is the rotation error in degrees, RMSE the translation root mean square error in $\times10^{-2}$, PA in percent, and CD in $\times10^{-3}$. Metric definitions follow GARF~\citep{li2025garf}; the evaluation setting and the baseline results follow CRAG~\citep{jiang2026crag}. Best in bold, second best underlined.}
\label{tab:partnext}
\begin{center}
\scriptsize
\setlength{\tabcolsep}{4pt}
\begin{tabular}{l cccc}
\toprule
\textbf{Method} & RE$\downarrow$ & RMSE$\downarrow$ & PA$\uparrow$ & CD$\downarrow$ \\
\midrule
GARF & 52.52 & 10.68 & 58.19 & \underline{3.10} \\
RPF & 54.99 & 29.54 & 46.17 & 10.46 \\
CRAG & \underline{45.12} & \underline{9.13} & \underline{65.89} & \textbf{2.40} \\
Assembler & 85.82 & 17.14 & 44.18 & 27.93 \\
\textbf{\method{} (ours)} & \textbf{41.12} & \textbf{5.24} & \textbf{67.93} & 6.37 \\
\bottomrule
\end{tabular}
\end{center}
\end{minipage}\hfill
\begin{minipage}[t]{0.47\linewidth}
\caption{Ablation study on the PartNet Chair subset. CD in $\times10^{-3}$; PA, CA, and SR in percent. The best result per column is in bold.}
\label{tab:ablation}
\begin{center}
\scriptsize
\setlength{\tabcolsep}{4pt}
\begin{tabular}{l cccc}
\toprule
\textbf{Variant} & CD$\downarrow$ & PA$\uparrow$ & CA$\uparrow$ & SR$\uparrow$ \\
\midrule
Full model & \textbf{6.21} & \textbf{71.42} & \textbf{53.87} & \textbf{41.40} \\
w/o summary block & 8.52 & 63.71 & 53.41 & 37.14 \\
MLP projector & 9.75 & 66.53 & 52.38 & 36.27 \\
w/o instruction & 8.56 & 65.29 & 51.93 & 34.78 \\
\bottomrule
\end{tabular}
\end{center}
\end{minipage}
\end{table}

\subsection{Generalization and Scalability}
\label{sec:analysis}

\paragraph{Zero-shot generalization to unseen categories.}
\label{sec:zeroshot}
To evaluate zero-shot generalization, we train \method{}, SPAFormer~\citep{xu2025spaformer}, 3DHPA~\citep{du_generative_2024}, and CFPA~\citep{zhang2025cfpa} only on Chair, Table, and Lamp, the PartNet protocol of Section~\ref{sec:setup}, and test them directly on three unseen categories, Bed, Door, and Storage Furniture, without fine-tuning (Table~\ref{tab:complex-categories}). The baselines generalize poorly: the highest baseline PA is 10.29\%, 15.78\%, and 9.13\% on the three categories (CFPA in each case), whereas \method{} reaches 31.03\%, 32.49\%, and 26.35\%, at least 16.7 points higher, together with the lowest CD and the lowest RMSE(R) on every category.

\begin{table}[t]
\caption{Zero-shot generalization to unseen PartNet categories; all methods are trained on Chair, Table, and Lamp only. CD in $\times10^{-3}$, PA in percent, RMSE(R) in degrees; best per column in bold.}
\label{tab:complex-categories}
\begin{center}
\footnotesize
\setlength{\tabcolsep}{2.4pt}
\begin{tabular}{l ccc ccc ccc}
\toprule
& \multicolumn{3}{c}{\textbf{Bed}} & \multicolumn{3}{c}{\textbf{Door}} & \multicolumn{3}{c}{\textbf{Storage Furniture}} \\
\cmidrule(lr){2-4}\cmidrule(lr){5-7}\cmidrule(lr){8-10}
\textbf{Method} & CD$\downarrow$ & PA$\uparrow$ & RMSE(R)$\downarrow$ & CD$\downarrow$ & PA$\uparrow$ & RMSE(R)$\downarrow$ & CD$\downarrow$ & PA$\uparrow$ & RMSE(R)$\downarrow$ \\
\midrule
SPAFormer~\citep{xu2025spaformer} & 50.24 & 5.81 & 81.04 & 49.23 & 6.63 & 83.56 & 48.47 & 4.21 & 87.12 \\
3DHPA~\citep{du_generative_2024} & 67.12 & 6.27 & 75.35 & 56.42 & 10.96 & 72.67 & 61.31 & 8.13 & 69.19 \\
CFPA~\citep{zhang2025cfpa} & 34.48 & 10.29 & 75.93 & 26.48 & 15.78 & 69.02 & 47.35 & 9.13 & 78.26 \\
\textbf{\method{} (ours)} & \textbf{18.45} & \textbf{31.03} & \textbf{40.76} & \textbf{21.15} & \textbf{32.49} & \textbf{53.71} & \textbf{24.14} & \textbf{26.35} & \textbf{55.78} \\
\bottomrule
\end{tabular}
\end{center}
\end{table}

\begin{figure}[t]
\begin{center}
\includegraphics[width=\linewidth]{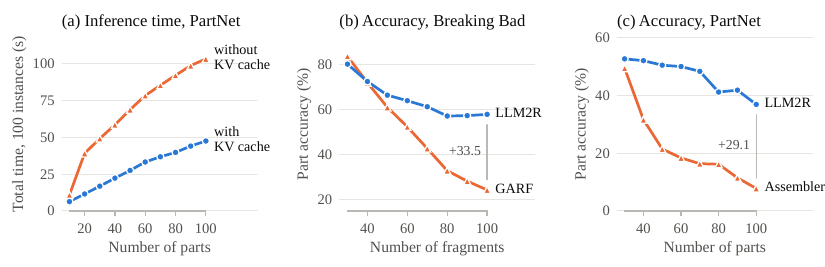}
\end{center}
\caption{Scaling with the number of parts, without retraining. (a) Total inference time over 100 PartNet instances per part count, with and without key-value cache reuse; (b) PA on Breaking Bad test instances with 30 to 100 fragments, against GARF; (c) PA on PartNet test instances with 30 to 100 parts, against Assembler.}
\label{fig:scaling}
\end{figure}

\paragraph{Scalability with the number of parts.}
\label{sec:scalability}
All poses are obtained from a single forward pass over $L$ tokens (Eq.~\eqref{eq:sequence}), and the summary block adds a constant $S$ tokens irrespective of $N$; with a context limit of 8{,}192 tokens and $K=64$, one sequence accommodates up to about 125 parts. Figure~\ref{fig:scaling} reports inference time and accuracy as a function of the part count, using the models of Section~\ref{sec:setup} without retraining. Figure~\ref{fig:scaling}(a) measures the total wall-clock time for 100 PartNet instances per part count when the part blocks are appended to the decoder one at a time. With the key-value cache reused across blocks, the time grows nearly linearly, from 6.33\,s at 10 parts (63\,ms per instance) to 47.54\,s at 100 parts, whereas recomputing the prefix for every block grows superlinearly to 103.37\,s. Figures~\ref{fig:scaling}(b) and (c) compare PA on test instances with 30 to 100 parts. On Breaking Bad, GARF is more accurate at 30 fragments (83.61\% against 80.13\%) but degrades to 24.33\% at 100 fragments, whereas \method{} stays above 57\% and reaches 57.85\% at 100 fragments, 33.5 points higher. On PartNet, \method{} and Assembler are close at 30 parts (52.57\% against 49.34\%), but Assembler collapses to 7.78\% at 100 parts while \method{} retains 36.83\%, an improvement of 29.1 points.

\subsection{Ablation Studies}
\label{sec:ablation}

\paragraph{Components.}
Table~\ref{tab:ablation} ablates the design of \method{} on the PartNet Chair subset. \emph{w/o summary block} reads each pose from the last token of its part block, i.e., a plain causal readout; \emph{MLP projector} replaces the Q-Former with a two-layer MLP applied to each patch descriptor; \emph{w/o instruction} removes the natural-language content of the prompt while keeping the chat template. Removing the summary block causes the largest drop in PA ($-7.7$ points) and raises CD by 2.31, whereas CA is almost unchanged, so the summary mainly improves the absolute placement of parts rather than their local connectivity. The MLP projector lowers PA by 4.9 points and raises CD by 3.5, so the learned queries yield a more useful part representation than a direct projection of the neighborhoods. Removing the instruction lowers PA by 6.1 points and SR by 6.6 points, the largest drop in SR. The geometric tokens, the summary block, and the pose head are identical in both variants, so the difference stems from the instruction tokens alone: the pretrained decoder makes use of the instruction when it processes the geometric tokens. This is the most direct evidence in our study that the language pretraining of the backbone, and not only its capacity, contributes to assembly accuracy.

\paragraph{Token budget.}
Table~\ref{tab:tokens} (left) varies the number of Q-Former queries $K$ with $S=K$ and all other settings unchanged. CD, PA, and SR improve monotonically with $K$, whereas CA stays within 0.2 points; the gains saturate beyond $K=64$, as doubling the budget to 128 improves PA by only 0.13 points and CD by 0.08 while doubling the number of geometric tokens.

\paragraph{Accuracy along the part sequence.}
Parts are presented in the dataset order and the model is not permutation invariant (Section~\ref{sec:summary}); no predicted pose is fed back into the decoder, so errors cannot accumulate through the predictions, but early parts see a shorter prefix. Table~\ref{tab:tokens} (right) reports the per-position PA of \method{} on 20-part Chair instances: PA is highest at the first two positions (83.4\% and 80.9\%), lies between 61.5\% and 70.9\% from position 5 to 17, and is lowest at positions 18 to 20 (56.4\% to 59.6\%). The decline is gradual, about 24 points over 20 positions, without a collapse at the tail. The first positions see the shortest prefix yet are the most accurate, so prefix length alone does not explain the trend once the summary states are available; we hypothesize that later positions are more often occupied by small or repeated parts whose placement is intrinsically ambiguous.

\begin{table}[t]
\caption{Left: PartNet Chair results for different numbers of geometric tokens per part $K$ ($S=K$); CD in $\times10^{-3}$, PA, CA, and SR in percent. Right: per-position PA (\%) on 20-part Chair instances, $n$ being the position in the input sequence.}
\label{tab:tokens}
\begin{center}
\scriptsize
\begin{minipage}[t]{0.4\linewidth}
\centering
\setlength{\tabcolsep}{4pt}
\begin{tabular}{c cccc}
\toprule
$K=S$ & CD$\downarrow$ & PA$\uparrow$ & CA$\uparrow$ & SR$\uparrow$ \\
\midrule
16 & 7.24 & 68.37 & 53.79 & 40.56 \\
32 & 6.96 & 70.88 & 53.70 & 41.02 \\
64 & 6.21 & 71.42 & 53.87 & 41.40 \\
128 & 6.13 & 71.55 & 53.81 & 41.67 \\
\bottomrule
\end{tabular}
\end{minipage}\hfill
\begin{minipage}[t]{0.58\linewidth}
\centering
\setlength{\tabcolsep}{2.5pt}
\begin{tabular}{l cccccccccc}
\toprule
$n$ & 1 & 2 & 3 & 4 & 5 & 6 & 7 & 8 & 9 & 10 \\
PA & 83.4 & 80.9 & 73.9 & 73.2 & 70.9 & 69.3 & 70.6 & 67.2 & 67.1 & 70.2 \\
\midrule
$n$ & 11 & 12 & 13 & 14 & 15 & 16 & 17 & 18 & 19 & 20 \\
PA & 63.6 & 65.8 & 62.5 & 61.7 & 61.7 & 61.5 & 62.3 & 56.4 & 58.1 & 59.6 \\
\bottomrule
\end{tabular}
\end{minipage}
\end{center}
\end{table}

\section{Conclusion}
\label{sec:conclusion}

We have presented \method{}, which casts 3D part assembly as causal sequence modeling over geometric part tokens with a LoRA-adapted pretrained decoder-only LLM. A learnable summary block appended after all part tokens aggregates the entire assembly within one forward pass, and a part--summary fusion head conditions every pose on this context at a constant token cost, so that one architecture handles semantic and geometric assembly without autoregressive decoding. \method{} attains the best part accuracy on PartNet and PartNeXt and the lowest translation error and inference time on Breaking Bad, where only the fracture-specialized GARF is more accurate; its advantage over the baselines grows with the number of parts, and it transfers zero-shot to unseen categories. Ablations show that the summary block contributes most to part accuracy and that the decoder makes use of the language instruction, so the pretrained backbone is used for more than its capacity.

\paragraph{Limitations and future work.}
\method{} is not invariant to the part order, and the context window of the decoder bounds the number of parts that can be processed at once (about 125 parts at $K=64$ with a limit of 8{,}192 tokens). It regresses a single pose per part, so it cannot express several plausible assemblies, and PCA-derived pose targets are ambiguous up to part symmetries, which the Chamfer objective only partly compensates. Its Chamfer distance lags behind hierarchical and fracture-specialized methods, which we attribute to the few parts it misplaces far from the surface. Order-invariant position encodings, larger backbones, and a lightweight geometric refinement of the predicted poses are natural next steps.

\subsection*{Reproducibility statement}
Section~\ref{sec:method} specifies the architecture, including the construction of the summary block, the causal information flow, the part--summary fusion head, the rotation parameterization, and all training objectives. Section~\ref{sec:setup} and Appendix~\ref{sec:implementation_appendix} list all implementation settings, including the Q-Former and pose-head architectures, the LoRA configuration, the optimizer, and the number of trainable parameters. Appendix~\ref{sec:data_processing} describes the construction of assembly instances from PartNeXt, the canonical-frame definition, and the derivation of pose targets. Source code and processed data manifests will be released upon publication.

\bibliography{main,related,related_add,intro,extra}
\bibliographystyle{llm2r_preprint}

\clearpage
\appendix
\noindent\textbf{Appendix overview.}
This appendix complements the main paper along three dimensions. Appendix~\ref{sec:data_processing} describes the construction of assembly instances, the canonical frames, and the derivation of pose targets. Appendix~\ref{sec:implementation_appendix} lists the implementation settings, including the fixed quaternion anchor set and the evaluation procedure. Appendix~\ref{sec:qualitative_appendix} provides additional qualitative results.

\section{Data Processing}
\label{sec:data_processing}

\subsection{Assembly Instances from PartNeXt}
\label{sec:partnext_data}
PartNeXt provides textured meshes with part masks defined over mesh faces. We adopt the split of CRAG~\citep{jiang2026crag}, with 15,563 training and 3,903 evaluation shapes that each contain between 2 and 20 parts, and train the PartNeXt model on its training split. For each shape, we read the GLB file, apply the node transformations of the scene graph, and gather the vertices of the faces assigned to each part mask into one point set per part; shapes without usable masks fall back to one part per scene primitive. The point sets of a shape are then normalized jointly, so that the center of the bounding box of the complete object lies at the origin and its longest side has unit length, and each part is stored as a vertex-only OBJ file.

\subsection{Canonical Frames and Pose Targets}
\label{sec:canonical}
Each part is associated with a canonical reference frame constructed from its own geometry by Principal Component Analysis (PCA), and the learning target of a part is the rigid transformation from this canonical frame to the frame of the assembled object. Given the vertex set of a part in the object frame,
\[
\mathcal{V}=\{\mathbf{p}_k\}_{k=1}^{M}, \qquad \mathbf{p}_k \in \mathbb{R}^3,
\]
we compute its centroid $\boldsymbol{\mu}=\frac{1}{M}\sum_{k}\mathbf{p}_k$ and the covariance matrix $\mathbf{\Sigma}=\frac{1}{M}\sum_{k}(\mathbf{p}_k-\boldsymbol{\mu})(\mathbf{p}_k-\boldsymbol{\mu})^\top$. The eigendecomposition $\mathbf{\Sigma}\mathbf{v}_j=\lambda_j\mathbf{v}_j$ with $\lambda_1\ge\lambda_2\ge\lambda_3\ge0$ yields the principal axes, and the third axis is flipped if necessary so that the frame is right-handed ($\det=+1$). The canonical rotation is then $\mathbf{R}_{\mathrm{can}}=[\mathbf{v}_1,\mathbf{v}_2,\mathbf{v}_3]\in\mathrm{SO}(3)$, and the rigid transformation that maps the part from its canonical frame, in which it is centered at the origin with axes aligned to the principal directions, to its placement in the object is
\[
\mathbf{T}_{\mathrm{can}\to\mathrm{obj}}=
\begin{bmatrix}
\mathbf{R}_{\mathrm{can}} & \boldsymbol{\mu} \\
\mathbf{0}^\top & 1
\end{bmatrix}
\in \mathrm{SE}(3).
\]
The supervision target of part $i$ is therefore $(R_i^\star,\bt_i^\star)=(\mathbf{R}_{\mathrm{can}},\boldsymbol{\mu})$, with $\bq_i^\star$ the unit quaternion of $\mathbf{R}_{\mathrm{can}}$ in scalar-first order and nonnegative scalar part. Parts with fewer than three vertices receive the identity rotation or a frame aligned with the segment joining their two vertices. This reformulation casts pose prediction as recovering the placement of a geometry-consistent canonical part rather than an absolute pose, so that the network can focus on inter-part spatial reasoning. Because the sign of each principal axis and the order of axes with similar eigenvalues are arbitrary, the canonical frame of a symmetric part is ambiguous, which is the source of the rotation-label ambiguity discussed in Section~\ref{sec:conclusion} and the motivation for the Chamfer objective.

\subsection{Input Coordinates and Data Splits}
\label{sec:input_frame}
The preprocessing above stores the vertices of every part in the normalized object frame and derives the pose target from those coordinates. Before tokenization, each part is expressed in its canonical frame, i.e., its vertices are centered at the centroid and rotated into the principal axes, so that the network observes only the intrinsic geometry of the part and predicts the transform $\mathbf{T}_{\mathrm{can}\to\mathrm{obj}}$ that returns it to the object frame. The point encoder consumes the vertex set directly; parts with fewer points than required by the grouping are padded by repeating randomly selected points. For PartNet and Breaking Bad, we adopt the official data splits provided by the respective dataset authors and the same canonical-frame construction.

\section{Implementation Details}
\label{sec:implementation_appendix}

\subsection{Model and Training Settings}
Table~\ref{tab:hparams} lists the settings of the main configuration. The point encoder follows the Point-BERT configuration with a 12-layer, 384-dimensional Transformer with 6 heads, a discrete vocabulary of 8{,}192 point tokens, and $G=64$ neighborhoods of 32 points obtained by farthest point sampling and $k$-nearest-neighbor grouping; parts with fewer points than required by the grouping are padded by repeating randomly selected points. The encoder is frozen and kept in evaluation mode for the first 80 epochs, so that its normalization statistics do not drift, and is fine-tuned end-to-end for the last 20 epochs. The Q-Former projects each patch descriptor with $\mathrm{Linear}(8192\to d)$ followed by layer normalization, refines $K=64$ learnable queries with $N_\ell=2$ pre-norm layers, each with 8-head self-attention among the queries, 8-head cross-attention to the patch features, and a feed-forward sublayer of width $4d$ with GELU, and applies a final layer normalization ($d=1024$); the ablated MLP projector is $\mathrm{Linear}(8192\to d)\to\mathrm{GELU}\to\mathrm{Linear}(d\to d)$. The pose head comprises the summary block $S^{(0)}\in\R^{64\times1024}$, the fusion layer $\mathrm{Linear}(2d\to d)$ with GELU, and two branches of the form $\mathrm{Linear}(d\to2048)\to\mathrm{GELU}\to\mathrm{Dropout}(0.05)\to\mathrm{Linear}(2048\to2048)\to\mathrm{GELU}\to\mathrm{Linear}(2048\to\cdot)$ with 32 outputs (followed by layer normalization) for rotation and 3 outputs for translation. LoRA is applied to the query, key, value, and output projections of the six full-attention layers of the backbone (24 adapted modules) with rank 32, scaling 64, and dropout 0.05. The summary block and the pose head are saved and restored together as one module. The instruction consists of a fixed system prompt (``You are a helpful assistant for furniture assembly.'') followed by a short user request for the 6D poses drawn from a pool of paraphrases; no text is generated or supervised.

\begin{table}[h]
\caption{Settings of the main configuration.}
\label{tab:hparams}
\begin{center}
\small
\begin{tabular}{p{4.1cm} p{8.6cm}}
\toprule
\textbf{Component} & \textbf{Setting} \\
\midrule
Language backbone & Qwen3.5-0.8B language model; 24 layers, $d=1024$; 3 gated linear-attention layers per full-attention layer \\
Point encoder & Point-BERT (ShapeNet pretrained); $G=64$ groups of 32 points; $C=8192$; frozen for epochs 1--80, fine-tuned for epochs 81--100 \\
Projector & Q-Former: $\mathrm{Linear}(8192\to1024)$+LN, $K=64$ learnable queries, $N_\ell=2$ layers (8-head self-attention, 8-head cross-attention, FFN of width 4096) \\
Tokens & $K=64$ per part; summary block $S=64$; context limit 8{,}192 tokens \\
Pose head & Fusion $\mathrm{Linear}(2048\to1024)$+GELU; two 3-layer MLPs of width 2{,}048; dropout 0.05 \\
Rotation output & Softmax mixture of 32 fixed quaternion anchors, normalized \\
LoRA & $r=32$, $\alpha=64$, dropout 0.05; q/k/v/o projections of attention layers \\
Trainable parameters & 42.2M (while the encoder is frozen) \\
Optimizer & AdamW, learning rate $10^{-3}$, $\beta=(0.9,0.95)$, weight decay 0; constant schedule \\
Precision & bfloat16 mixed precision \\
Checkpoint selection & Lowest validation loss, evaluated after every epoch \\
Epochs, batch size, hardware & 100 epochs; total batch size 32 (4 per GPU); 8 NVIDIA H200 GPUs; about 10 hours \\
Loss weights & $\lambda_R=\lambda_t=\lambda_{\mathrm{CD}}=\lambda_{\mathrm{rep}}=1$ \\
Prompt layout & Geometric tokens and summary block inside the user message, after the instruction text \\
\bottomrule
\end{tabular}
\end{center}
\end{table}

\subsection{Quaternion Anchor Set}
\label{sec:anchors}
The rotation head mixes $N_a=32$ fixed unit quaternions $\mathbf{c}_m=(w,x,y,z)$. Eight anchors are the signed unit basis vectors $\pm(1,0,0,0)$, $\pm(0,1,0,0)$, $\pm(0,0,1,0)$, and $\pm(0,0,0,1)$, i.e., the identity and the $180^\circ$ rotations about the three coordinate axes, each with both quaternion signs. The remaining 24 anchors have exactly two nonzero entries equal to $\pm\tfrac{1}{\sqrt2}$: those combining the scalar part with one imaginary part represent the $\pm90^\circ$ rotations about the coordinate axes, and those combining two imaginary parts represent the $180^\circ$ rotations about the six axis bisectors, again with both signs. The set thus contains 16 distinct rotations of the octahedral group, each represented twice, and positively spans $\R^4$, so that the normalized convex combination in Eq.~\eqref{eq:rotation} can express any rotation.

\subsection{Evaluation Procedure}
\label{sec:metric_audit}
For every evaluated part, the evaluation script records the geodesic rotation error between the predicted and the target quaternion, the Euclidean translation error, and the squared Chamfer distance between the part transformed by the predicted and by the target pose, computed on at most 2{,}048 randomly selected points, together with the indicator of whether this distance falls below the PA threshold of 0.01. Inference time is measured with device synchronization around the model call and divided by the number of evaluated instances. These per-part quantities are aggregated into RMSE(R), RMSE(T), PA (threshold 0.01), and whole-object CD following the protocol of GARF, and into CD, PA, CA, and SR on PartNet following the protocol of Assembler, as described in Section~\ref{sec:setup}.

\section{Additional Qualitative Results}
\label{sec:qualitative_appendix}
\begin{figure}[p]
\begin{center}
\includegraphics[width=0.92\linewidth]{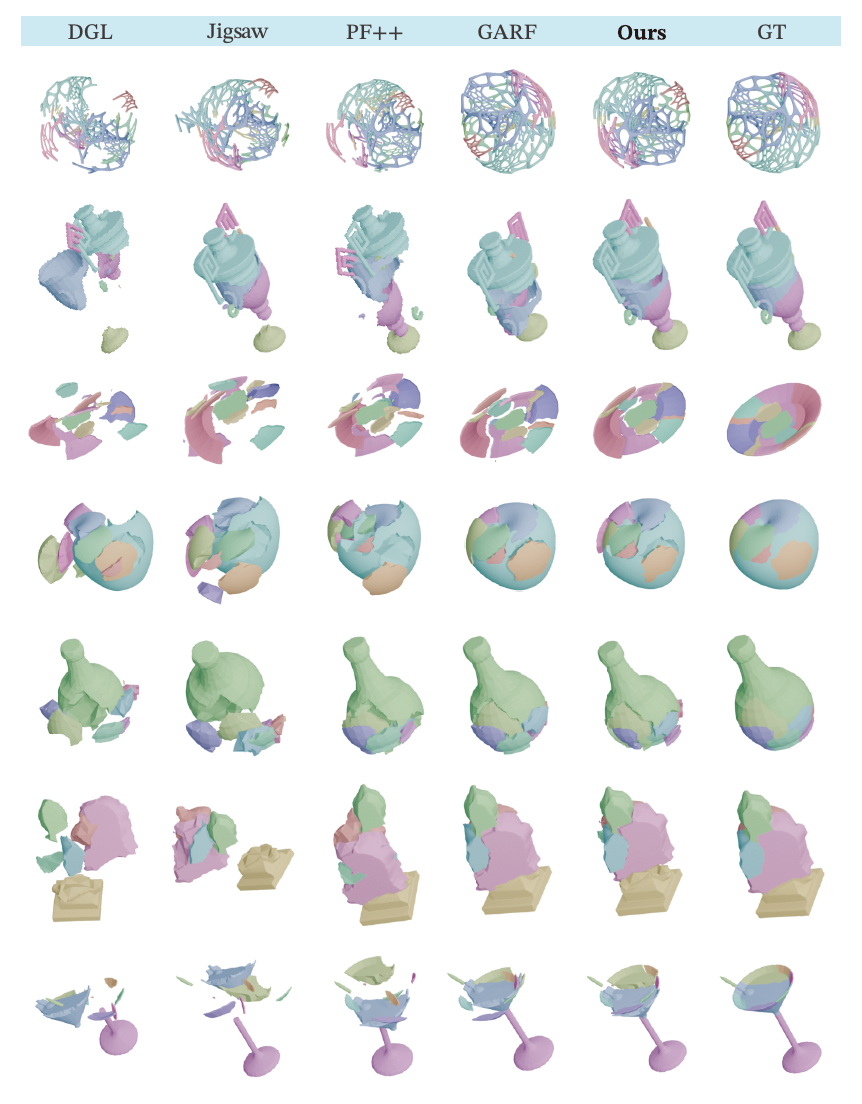}
\end{center}
\caption{Qualitative comparison on Breaking Bad. Columns show DGL, Jigsaw, PF++, GARF, \method{} (ours), and the ground truth. \method{} recovers globally coherent assemblies whose fragments are placed close to their ground-truth locations, whereas the baselines more often leave fragments displaced or scattered.}
\label{fig:qual}
\end{figure}

\begin{figure}[p]
\begin{center}
\includegraphics[width=0.92\linewidth]{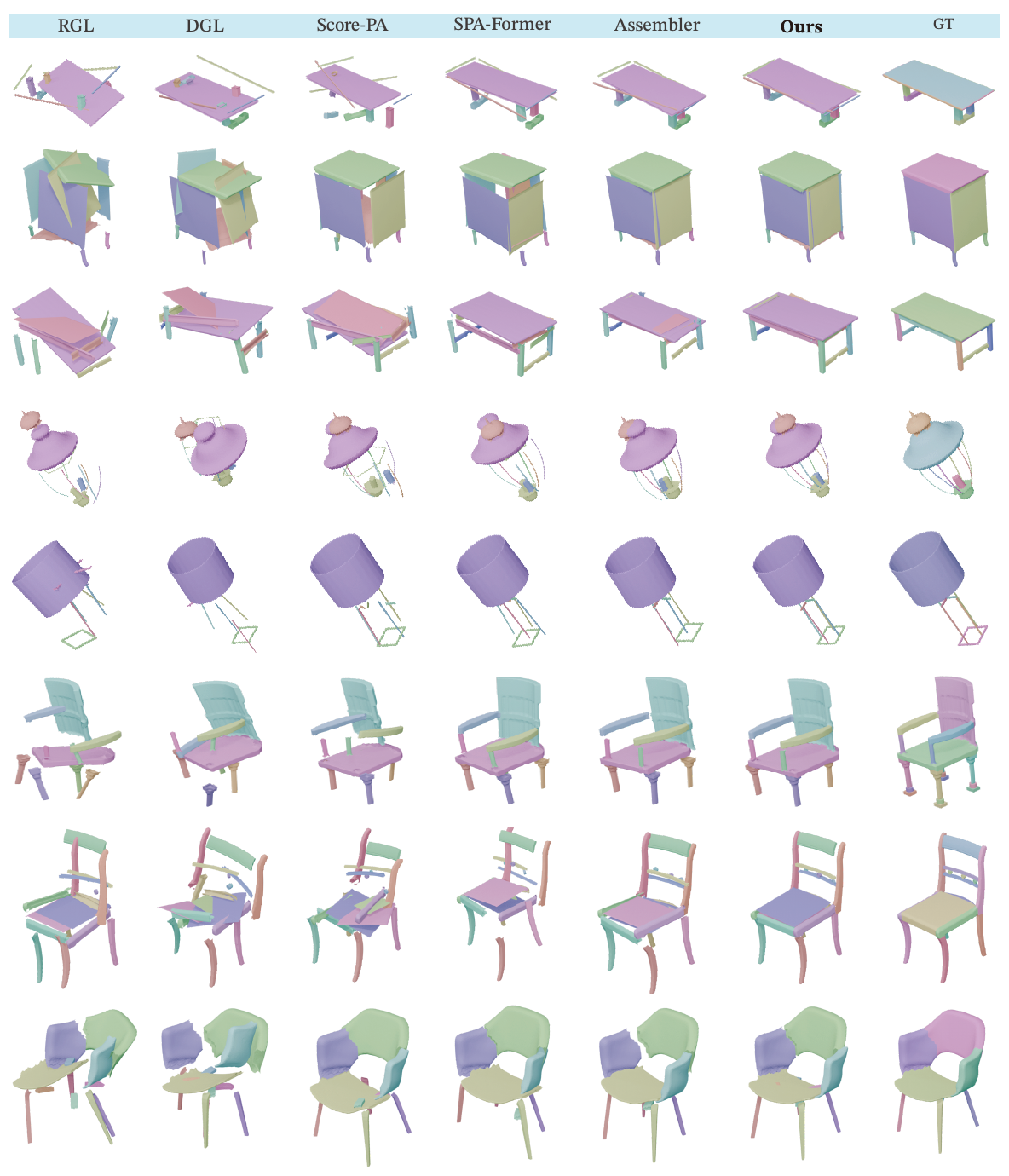}
\end{center}
\caption{Qualitative comparison on PartNet Table, Lamp, and Chair instances. Columns show RGL, DGL, Score-PA, SPAFormer, Assembler, \method{} (ours), and the ground truth. \method{} places parts with accurate poses across diverse object structures, producing complete and well-aligned assemblies where several baselines exhibit misaligned or detached parts.}
\label{fig:qual_partnet}
\end{figure}

\begin{figure}[t]
\begin{center}
\includegraphics[width=0.92\linewidth]{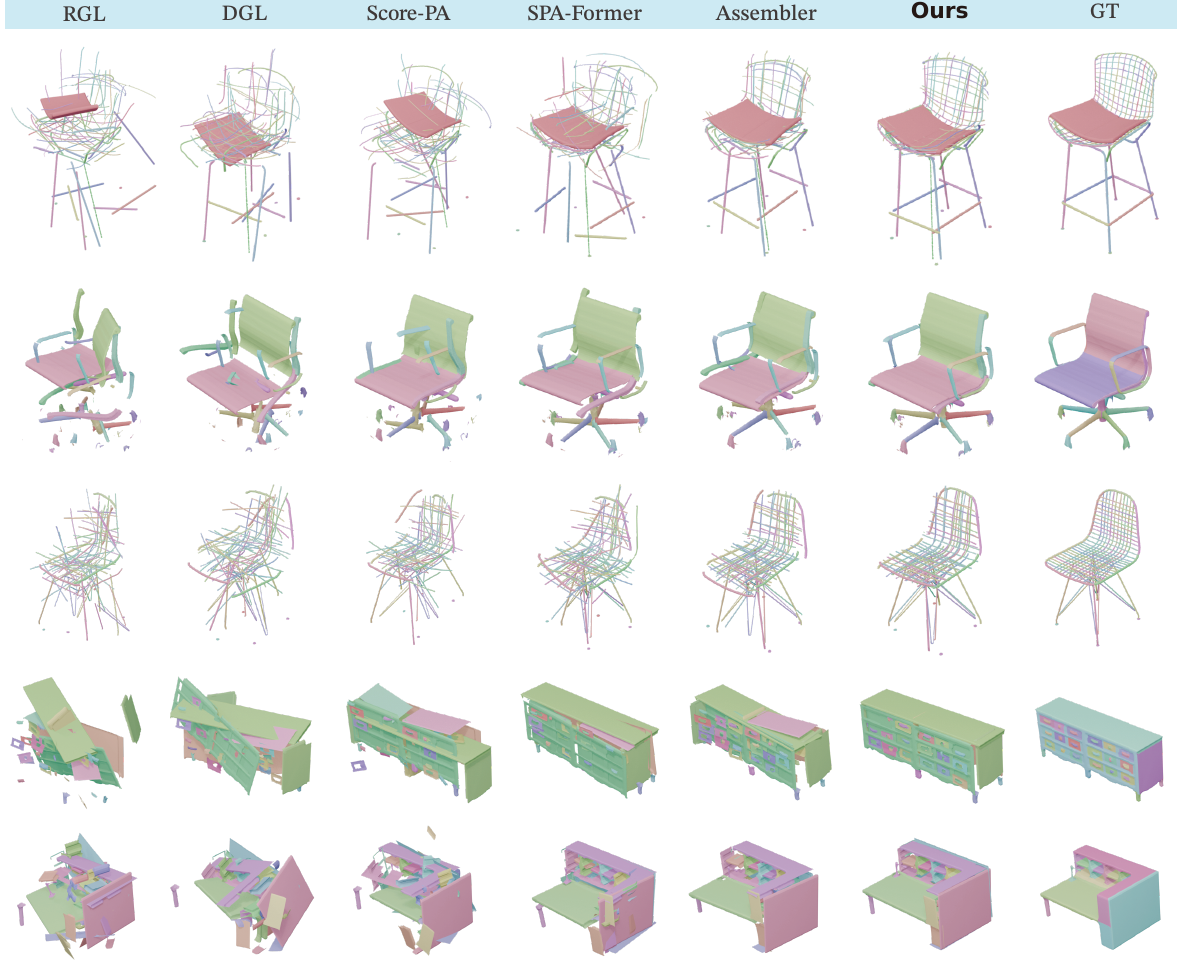}
\end{center}
\caption{Qualitative comparison on PartNet instances with a large number of parts, such as wire-frame chairs and storage furniture with many drawers. Columns follow Figure~\ref{fig:qual_partnet}. \method{} produces coherent assemblies in which the many thin and repeated parts are placed consistently, whereas the baselines frequently yield collapsed or misaligned structures.}
\label{fig:qual_partnet_many}
\end{figure}

Figures~\ref{fig:qual}--\ref{fig:qual_partnet_many} show qualitative comparisons produced by the summary-conditioned model. On Breaking Bad (Figure~\ref{fig:qual}), \method{} is compared with DGL, Jigsaw, PF++, and GARF; on PartNet (Figure~\ref{fig:qual_partnet}), with RGL, DGL, Score-PA, SPAFormer, and Assembler on Table, Lamp, and Chair instances; and Figure~\ref{fig:qual_partnet_many} shows PartNet instances with a large number of parts. The ground truth is shown in the last column of each figure.

\end{document}